# Unlocking Futures: A Natural Language Driven Career Prediction System for Computer Science and Software Engineering Students

Sakir Hossain Faruque[a], Sharun Akter Khushbu[a], Sharmin Akter[a]

[a]*Department of Computer Science and Engineering, Daffodil International University, Dhaka, Bangladesh*

**Abstract**

A career is a crucial aspect for any person to fulfill their desires through hard work. During their studies, students cannot find the best career suggestions unless they receive meaningful guidance tailored to their skills. Therefore, we developed an AI-assisted model for early prediction to provide better career suggestions. Although the task is difficult, proper guidance can make it easier. Effective career guidance requires understanding a student's academic skills, interests, and skill-related activities. In this research, we collected essential information from Computer Science (CS) and Software Engineering (SWE) students to train a machine learning (ML) model that predicts career paths based on students' career-related information. To adequately train the models, we applied Natural Language Processing (NLP) techniques and completed dataset pre-processing. For comparative analysis, we utilized multiple classification ML algorithms and deep learning (DL) algorithms. This study contributes valuable insights to educational advising by providing specific career suggestions based on the unique features of CS and SWE students. Additionally, the research helps individual CS and SWE students find suitable jobs that match their skills, interests, and skill-related activities.

## 1. Introduction

The rapid growth of the IT industry has peaked students' interest in CS and SWE. With so many employment choices, many graduates in various professions may struggle to choose the ideal career path. This issue stems from a mismatch between labor market needs and students' skill sets, which can result in workplace discontent, poor performance, and stress (Madhan & Reddy, 2021a). Despite educators' best attempts to direct students to meaningful work prospects, the complex interplay of factors impacting career choices remains a considerable obstacle.

In light of this critical issue, we investigate one aspect of ML, NLP, to help undergraduate CS and SWE students make educated career choices. The previous study *(2018 Fourth*

---

*Corresponding authors.
Email address: sakir15-3862@diu.edu.bd (Sakir Hossain Faruque) sharun.cse@diu.edu.bd (Sharun Akter Khushbu)

*International Conference on Computing Communication Control and Automation (ICCUBEA),* n.d.; Kabakus & Senturk, 2020; Lent et al., 2008; Madhan & Reddy, 2021a) discovered that academic performance had a significant influence on work decisions, demonstrating a link between proficiency in certain courses and interest in associated industries. Furthermore, these studies underline the importance of individual abilities and interests in choosing career paths, with people naturally gravitating toward occupations where they can succeed and have a strong desire for continuing growth and advancement.

Building on this basis, our study investigates how ML techniques, namely NLP, may aid undergraduate CS and SWE students in making educated career decisions. The current body of research (*2018 Fourth International Conference on Computing Communication Control and Automation (ICCUBEA),* n.d.; Kabakus & Senturk, 2020; Lent et al., 2008; Madhan & Reddy, 2021a) has demonstrated the importance of academic accomplishment in influencing career choices, indicating a relationship between knowledge of certain courses and a willingness to fulfill professional responsibilities. Furthermore, these studies emphasize the importance of students' talents and interests in determining their future routes, with a significant preference for careers that match their capabilities and aspirations. Following a rigorous collection and evaluation of career-related data from undergraduate students, we discovered significant determinants impacting professional decisions. We utilized these data to create ML models powered by NLP that accurately predicted career paths. Now the question can be appear how this research contributes in real life.

- This enables students to confidently traverse the vast terrain of job choices, addressing the inherent problems of professional decision-making.

- Our findings broaden the use of NLP-enhanced career models beyond students to include academic advisers, career counselors, and industry experts, who may use these actionable insights to direct students along meaningful and profitable career paths.

- We also aim to understand the dynamics of career decision-making among undergraduate CS and SWE students through Google Forms. The data may refer to their future career paths, courses, skills, and related work.

Although there are overlaps related to the career paths and the titles of the skills, especially when it comes to computer science, such outcomes are not significant. Our objective is to understand the students' motivations and external influences that shape their career objectives. Our ambition is to develop NLP-driven career advising models that are better suited to the students' needs. In addition, we aim to discover novel techniques for addressing CP challenges using NLP technology, paying special attention to scalability, generalizability, and feature selection processes. Our study is comprehensive, and it equips students with the necessary knowledge and tools to manage their careers effectively in the highly dynamic IT industry.

## 2. Related works

The major goal of this article is to predict the career paths of CS and SWE students based on their attire. Existing research is analyzed, indicating a gap in comprehending the way students' academic and activity choices link to their job prospects. A comprehensive review of several studies reveals that only a limited number of research focus on forecasting the careers of undergraduate computer science students. Here, we aim to discuss the current studies that researchers have conducted.

2.1. Literature review

Ade and Deshmukh (Ade & Deshmukh, 2015) suggested an incremental ensemble learning approach for forecasting student career paths. They used voting to merge three classifiers (Naïve Bayes (NB), K-Star, and SVM). From these three classifiers, the accuracy percentage was respectively 89.6%, 89.2%, and 89.2%. They highlighted the importance of considering interest, talent, and projected growth in career choice. Mandalpu and Gong (Mandalapu & Gong, n.d.) discussed STEM and non-STEM career choices, which were two broad categories. STEM stands for Science, Technology, Engineering, and Mathematics, while Non-STEM refers to all other career fields outside of these four areas. They organized a survey on US middle school students where they provided career choices of 591 students, whereas 466 students belonged to NON-STEM and the remaining students belonged to STEM fields. Lent et al. (Lent et al., 2008) targeted to explore the role of self-efficacy, expectations, interests, social supports, and barriers in every student's academic and career choices. The research was conducted on 1208 students and used models such as the measurement model and structural model to analyze students' data. The measurement model was used to build the relationship between students' goals, supports, and barriers, whereas the structural model predicted educational and career goals based on self-efficacy, expectations, interests, supports, and barriers. The paper by Rangnekar et al. (*2018 Fourth International Conference on Computing Communication Control and Automation (ICCUBEA),* n.d.) studied students' aptitude, personality, and background, and the dataset was made to access these three basic principles. The dataset of this paper contained 2000 tuples, which included four subjects (Science, Mathematics, English, and Logical Reasoning) and career labels to identify suitable career choices for students. After completing all the preprocessing steps, the authors took 80% of the data for training and 20% of the data for testing from the aptitude and student information section, which was slightly different for the personality section (90% training and 10% testing). There were five ML algorithms used: K-nearest neighbors (KNN), Stochastic Gradient Descent (SGD), Logistic Regression (LR), Random Forest (RF), and Adaptive Boosting (AB). KNN and SGD provided high accuracy for career mapping, while LR and AB provided high accuracy for student background data. Casuat et al. (Casuat, 2020) investigated the high unemployment rate and showed that there was a mismatch between university-provided curriculum and job market demand. The research predicted employability based on 3000 entries and 12 features using ML algorithms such as SVM, DT, NB, and KNN. Among them, SVM achieved the highest accuracy of 92.22%, while DT achieved the lowest accuracy at only 56.36%. Shankdhar et al. (Shankhdhar et al., 2020) showed that choosing a career was difficult for college students who lacked self-awareness about their skills and interests. The research data were based on several attributes such as gender, nationality, age, attendance, participation in extracurricular activities, attentiveness in class, and the number of online courses taken. The authors claimed that the Decision Tree algorithm was the best to predict career but did not mention the accuracy percentage. Vignesh et al. (Vignesh et al., 2021) addressed the issue that students often face confusion in choosing the right career path after secondary education. The authors suggested a CP system that takes input from students' skills to help them choose a preferable career. The proposed system comprised three modules. In the Prediction Module, authors used both supervised and unsupervised learning. Under supervised learning, KNN and other algorithms were employed for classifying the target values, with KNN achieving approximately 94% accuracy. K-Means Clustering (KMC), an unsupervised learning algorithm, was used to cluster observations and to recommend departments based on students' performance. Sobnath et al. (Sobnath et al., 2020) mainly focused on the challenges faced by disabled students in choosing the right career and used ML and big data technologies for career guidance. The research worked on 270,934 students' records and applied missing value handling, label encoding, one-hot encoding, and other related activities to preprocess the dataset. Authors used ML algorithms such as LR, Linear Discriminant, DT Classifier, and Gaussian NB (GNB), among which LR and DT Classifier achieved the highest accuracy of 96% in predicting the Standard Occupation Classification (SOC).

Reddy (Madhan & Reddy, 2021a) applied XG Boost and a DT to predict the right career based on the student's skill. They collected data from students and underwent preprocessing steps to gain meaningful information. In the results section, they discussed how they predicted careers for undergraduate Engineering students (CS), but they did not show any accuracy percentage of the algorithms they used. Additionally, they did not discuss the confusion matrix and other elements needed to analyze the performance of the system. Kabakus and Senturk (Kabakus & Senturk, 2020) explained CS final-year students' project choices and how this impacts their professional careers. They focused on GitHub and IEEE to find project topics of the students using Python libraries PyGithub and IEEE Xplore. In this paper, a total of 14 ML algorithms were used to predict project topics and increase the accuracy level. Among all the algorithms, Bayes Net (BN), Kleene Star (KS), and J48 gave the best accuracy, which was 100%. Sinha et al. (Sinha & Singh, n.d.) demonstrated how ML algorithms can be much more effective in predicting careers. In the paper, the authors discussed the CP factors, data processing system (One hot encoding), and some algorithms (SVM, DT, XG Boost, etc.), but they did not show any accuracy percentage or prediction model, which are important steps to analyze the performance of the ML algorithms.

2.2. Comparison between existing works on CP

Table 1: Prior knowledge of the study of CP systems where reviewers show performance on Structured Supervised Learning (SSL):LR, Linear Discriminant (LD), DT Classifier, GNB, XG Boost, SVM, NB, KS, Gradient Boosted Tree (GBT), RF, RT, KNN, SGD, SGD, BN, Multinomial Naïve Bayes (MNB), J48, AdaBoost (AdaB), Stacking, and Bagging, and DANN (Deep Adversarial Neural Network): AutoMLP, Deep Neural Network (DNN), Multilayer Perceptron (MLP)

| SL No | Author Name | Used Algorithm | Best Accuracy with Algorithm |
|---|---|---|---|
| 1 | Roshani Ade and P. R. Deshmukh (Ade & Deshmukh, 2015) | SSL | NB = 89.6% |
| 2 | Varun Mandalapu and Jiaqi Gong (Mandalapu & Gong, n.d.) | SSL, DANN | GBT= 98.83% |
| 3 | V Madhan Mohan Reddy (Madhan & Reddy, 2021a) | SSL | DT has better performance |
| 4 | Rucha Hemant Rangnekar, Khyati Pradeep | SSL | Aptitude: |

| | Suratwala, Sanjana Krishna and Dr. Sudhir Dhage (2018 Fourth International Conference on Computing Communication Control and Automation (ICCUBEA), n.d.) | | SGD = 81.035% Personality (Myers Briggs classification): KNN = 81% Student Background Data: AB = 60.09% |
|---|---|---|---|
| 5 | Abdullah Talha Kabakus and Arafat Senturk (Kabakus & Senturk, 2020) | SSL, DANN | BN = 100%, KS = 100%, J48 = 100% |
| 6 | Cherry D. Casuat, Enrique D. Festijo and Alvin Sarraga Alon (Casuat, 2020) | SSL | SVM = 92.22% |
| 7 | Ashutosh Shankhdhar, Akash Agrawal, Deepak Sharma, Suryansh Chaturvedi and Mukesh Pushkarna (Shankhdhar et al., 2020) | SSL | DT has the best accuracy |
| 8 | Vignesh S, Shivani Priyanka C, Shree Manju H, Mythili K (Vignesh et al., 2021) | SSL | KNN = 94.10% |
| 9 | Ankita Sinha, Garima and Ashish Singh (Sinha & Singh, n.d.) | SSL | No accuracy found |
| 10 | Drishty Sobnath, Tobiasz Kaduk, Ikram Ur Rehman, and Olufemi Isiaq (Sobnath et al., 2020) | SSL | LR = 96% DT Classifier = 96% |

2.3. Gap analysis and finding according to prior study

Table 2: Finding of gap in the computer science students' CP system

| SL No | Research Work | Literature Review | Research Gap |
|---|---|---|---|
| 01 | Implemented Deep Learning | It is noticed that only a few papers have implemented deep learning in their CP research. (2018 Fourth International Conference on Computing Communication Control and Automation (ICCUBEA), n.d.; Mandalapu & Gong, n.d.) | As the research is working on multiple attributes of student information so deep learning implementation is a crucial need. |
| 02 | Model Accuracy | Some research works did not mention model accuracy but they used some ML algorithms. (Casuat, 2020; Madhan & | To know about model accuracy is very important so that we can justify our research progress. |

| 03 | Used Algorithm | It is seen that some papers have implemented only two to three algorithms and declared the output. (Ade & Deshmukh, 2015; Casuat, 2020; Madhan & Reddy, 2021a; Shankhdhar et al., 2020; Vignesh et al., 2021) | Only using two to three algorithms is not enough for research. Thus, we cannot properly compare with other algorithms. |
|---|---|---|---|
| 04 | Attributes Used in Dataset | These papers did not mention any specific attributes used in their dataset for the research project. (Madhan & Reddy, 2021a; Mandalapu & Gong, n.d.) | For students' CP we need to analyze various attributes such as students' exam marks, skills, interests, financial condition, etc. Mentioning attributes is very important for proper research work. |

## 3. Methods

The proposed methodology is about the overall research progress to develop a CP system for CS and SWE students' based on their field of skills, interests, and skill-based related works.

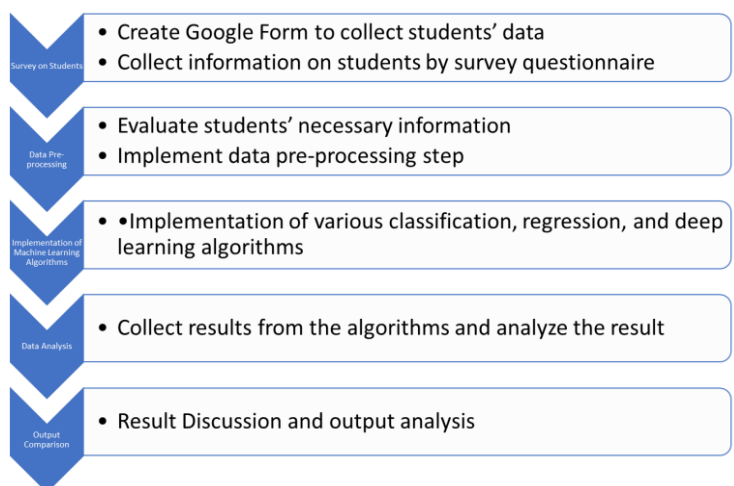

Figure 1: Methodology of extracting knowledge step by step

3.1. Survey questions on CP toward students' opinion

A Google Form was created to collect students' career information. In Table 3 we have shown the survey questionnaire with the necessary information.

Table 3: Survey related questions

| SL | Survey Questions | Key Fields of Students Opinion |
|---|---|---|

| NO | | |
|---|---|---|
| 01 | Student's Semester | 1st semester, 2nd semester, …, 12th semester |
| 02 | In the future, which field are you most interested in pursuing? | Software Development, Data Analysis, Higher Study, Researcher, …and others |
| 03 | If you choose a career as a researcher in which field do you want to research? (Depends on question 2) | Data Science, Artificial Intelligence, Human Computer Interaction, …, and Others |
| 04 | If you do higher study in which field you will target? (Depends on question 2) | Data Science and Analytics, Cybersecurity, Human Computer Interaction, …, and others |
| 05 | What do you think which CS core courses are the most important for your future? | Programming and Problem Solving, Data Structure, Database Management System, … |
| 06 | Out of academic study which skills do you think are most important to build up your career? (Depends on question 2) | Web Development, Mobile App Development, Artificial Intelligence (AI), Machine Learning, …, and others |
| 05 | Have you ever engaged in any activities that have contributed to your career? | Yes or No |
| 06 | If yes, in which field you have contributed to build your career? (Optional) | Research on Artificial Intelligence and Machine learning, Research on Natural Language Processing, …, and other |

Students' data were collected from various top universities in Bangladesh. We maintained a snowball process to collect data. The record of the students' information was stored in the Google Excel sheet.

3.2. Data analysis

After collecting the data, it was determined that around 19% of students picked software development as their career route, followed by 13% for web development and another 13% for cybersecurity. Approximately 10% exhibited interest in artificial intelligence, with 8% pursuing higher education. The remaining 53% of students picked other fields, each with a modest percentage Fig. 2.

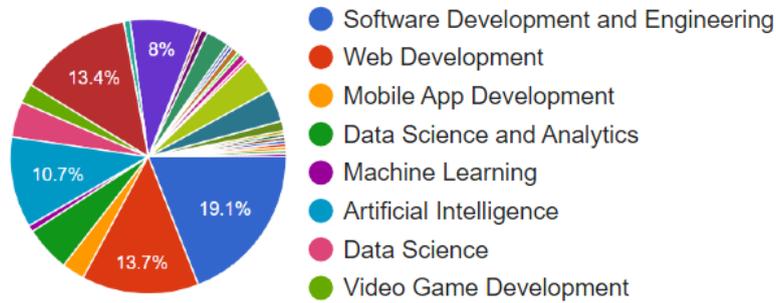

Figure 2: Student career field selection percentage

3.3. Feature engineering on CP

Data preparation involves using strategies to eliminate inconsistencies and noise from database inputs. These procedures improve the accuracy of the models (Agarwal, 2014). As the data is raw, it is necessary to preprocess it for better results. We followed various steps to preprocess the data.

Manual preprocessing

After collecting raw data, we found that students selected various fields they wanted to pursue for their careers. It was found that some fields were rarely chosen. So, we categorized those fields under some master fields. To categorize the fields under a master's field, we matched core skills and pieces of knowledge needed for the recommended fields. Similar skill and knowledge-based fields were kept under one master field. In Table 4 we have shown the master field and the field under those master fields.

Table 4: Master fields, students' chosen fields, and required skills for the chosen fields Mobile App Development (MAD), Video Game Development (VGD), Web Development (WD), Database Development (DD), Problem-Solving and Analysis (PSA), Critical Thinking (CT), Cloud Computing (CC), Project Management (PM), Interpersonal Skills (IS), Network Security (NS), System Design (SD), API Knowledge (AK), CyberSecurity (CS), Information Security Analysts (ISA), Cloud computing and DevOps (CCD), Tech Policy and Advocacy (TPA), Data Analysis and Visualization (DAV), Network Topologies (NT), Operating Systems and Networking (OSN), Data Engineering (DE), Robotics and Automation (RA), Internet of Things (IoT), Big Data Technologies (BDT), Feature Engineering (FE), Statistical Analysis (SA), Robotics Knowledge (RK), Software Development and Engineering (SDE), Consulting and IT Management (CIM), System Analyst (SysA), User Experience (UX) and User Interface (UI) Design (UI / UX), Graphic Design (GD), Video Editing (VE), Video Editing and Animation (VEA), Data Science (DS), Adaptability (ADP), Communication (COM), Cryptograph (CYP), Project Manager (PMG)

| Master Field | Students Suggested Field | Required Skill |
| --- | --- | --- |
| Development (DEV) | MAD, VGD, WD | WD, MAD, VGD, DD, PSA, CT, ADP, CC, PM, COM, IS, NS, SD, AK |
| Security(SEC) | CS, ISA, CCD, TPA | CS, DAV, NS, CT, AI and ML, CC, NT, ADP, PSA, OSN, CYP, DE, SD, COM, IS |
| AI | AI, RA, IoT, ML | PSA, DE, AI, and ML, DAV, CT, DD, BDT, FE, SA, SD, CC, RK, PM, IoT, COM, IS, ADP, CT |
| SDE | SDE, PMG, CIM, SysA | WD, MAD, AI and ML, PSA, CS, OSN, BDT, PM, DD, DAV, DE, SD, CC, COM, IS, ADP, CT |
| UI / UX | UI / UX, GD, VE | UI / UX Knowledge, GD, Working with Technology and New Idea Generate, VEA |
| DS | DS | PSA, DE, AI and ML, DAV, CT, DD, COM, BDT, FE, SA, SD, CC, CT, IS, ADP |

Handling garbage value

The database or data warehouse stores raw data that may be inconsistent and require cleaning and integration (Agarwal, 2014). It was noticed that many students did not properly read the questions on the form and provided skills and interests that did not match their chosen careers. If there was an extensive mismatch between the chosen and other related fields, then we removed those data. Sometimes, we made slight adjustments to align their career with other fields. Although it was a challenging task, we performed this activity for every row.

Eliminate redundant columns

The "Courses" and "Semester" categories were removed from the dataset due to a perceived lack of relevance to students' career options. Students were given a varied selection of courses to choose from autonomously, rendering the "Courses" section unnecessary. The omission of these columns was intended to allow students to choose career-related abilities and interests from the existing options. The "Semester" column was used to manage garbage values, although it did not affect the training data when applying ML techniques.

Extracted column from primary dataset

As the data was text data, it was necessary to apply NLP for other preprocessing steps and obtain the CP output. To apply the NLP library, we needed to combine the input fields into one field and keep the master fields as output. Here, the input field x is Skill, and the output field y is Career.

Stop words removal and lowercasing

When preprocessing text data, certain words lack significant value and instead contribute redundancies that can decrease model performance (Alshanik et al., 2020). In NLTK library there are various stop words such as a, the, or etc. Here, we removed stop words and made the words lowercase using the NLTK library.

Label encoding

Label Encoding is a technique for converting category columns into numerical ones so that they may be fitted by ML models that only accept numerical input. It is a critical pre-processing step when data is pre-processed. The Label Encoder approach was applied to label the master fields.

Table 5: Master field liberalization

| Label | Master Field |
|-------|--------------|
| 0 | AI |
| 1 | DS |
| 2 | DEV |
| 3 | SEC |
| 4 | SDE |
| 5 | UI / UX |

Text vectorization

Vectorization is a method in computer science and mathematics in which operations are performed on arrays (vectors) of data rather than individual pieces. It is a fundamental concept in numerical computing that significantly contributes to the efficiency of algorithms and calculations, particularly in data science, ML, and scientific computing. Text vectorization approaches can significantly improve classification accuracy (Krzeszewska et al., 2022). This method can enhance classification accuracy (Krzeszewska et al., 2022). Vectorization was used to determine the number of times a word occurred in a certain field.

Data train test split

Train-test split is a critical component of data science. The data was divided so that 80% of each master field was designated for training and the remaining 20% was retained for testing.

Data shuffling

In ML, the data shuffling technique allows organizations to share and analyze data while minimizing disclosure risk (Muralidhar & Sarathy, 2006). It is significant because it adds randomness without removing any underlying order or patterns in the collection. When data is presented in a specified order, such as all instances from one class followed by another, biased learning can occur during the training phase, which is a serious problem. Shuffling the data guarantees that the model meets a varied and random collection of cases during each training cycle, preventing it from learning patterns based on the data's sequence. Therefore, this practice helps in building a more robust and generalizable model.

3.4. Structured supervised Learning on CP system

ML is the study of utilizing computers to replicate how individuals learn and develop self-improvement techniques for gaining new knowledge and skills. It involves building models and algorithms that allow systems to discover current facts and constantly enhance performance (Li et al., 2009). In this phase, we will discuss implemented ML algorithms. So, the question can arise as to why we need to apply ML algorithms in our research as we have been researching CS and SWE students' CP so it is necessary to learn machines to predict the career of any student based on his or her skill, interest, and skill-based work. For this purpose, we used several ML algorithms to analyze the comparison of prediction.

DT

A supervised ML algorithm that is for both classification and regression tasks. It is similar to a flowchart where core nodes are represented by rectangles and leaf nodes by ovals (Priyam et al., 2013). Decision Tree learning is a highly effective technique due to its ease of use, comprehensibility, lack of factors, and ability to deal with mixed-type data (Su & Zhang, n.d.). As our dataset is already clean and pre-processed DT gives an average performance.

SVM

It is a supervised ML algorithm that uses a separating hyperplane to classify data points into different categories, maximizing the margin between classes for effective linear binary classification (Kumar & Jambheshwar, 2011; Support Vector Machines for Natural Language Processing, n.d.). In our pre-processed dataset it gives better performance.

LR

LR is a statistical approach used in ML to classify binary and multiclass data. It does quite well on the entire dataset.

KNN

It is a supervised ML method that can do both classification and regression tasks. The K-NN technique is a case-based learning approach that reviews all training data to identify the class with the most votes using k-nearest neighbors (Ariwa et al., n.d.). It gives an average performance for our dataset.

NB

NB is a probabilistic ML method that relies on Bayes' theorem (Lakhotia & Bresson, 2018). The word "naïve" signifies the idea that the existence or missing of one aspect does not affect the presence or lack of other traits. The Multinomial Naive Bayes Classifier applies the naive Bayes method to multinomially distributed data and is widely used for the categorization of texts (Lakhotia & Bresson, 2018). It gives an average performance on the dataset.

3.5. DANN on CP system

Deep learning (DL), a key technology of the Fourth Industrial Revolution based on artificial neural networks (ANN), is extensively used in a variety of computer applications due to its data-driven learning capabilities (Sarker, 2021). In this work, we used DL classifiers to improve the accuracy of student career predictions.

Convolutional Neural Network (CNN):

It is a multi-layer neural network (NN) that includes convolution and pooling layers. Each layer is made up of many two-dimensional planes, each with separate neurons. CNNs provide an important contribution to text categorization by automatically extracting significant features from text input, eliminating the need for human feature extraction (IEEE Staff, 2017).

MLP

A multilayer perceptron, also known as an MLP, is an ANN that is composed of many layers of connected nodes or neurons that follow the feedforward NN paradigm. It uses hidden layers and the backpropagation method for supervised learning, altering the weights of neural connections to reduce error. MLPs are often used in multiclass classification applications, with the softmax activation function in the output layer. The network is trained using gradient descent, which updates weights depending on backpropagation gradients (Kamath et al., 2018).

Long Short-Term Memory (LSTM)

It is a recurrent neural network (RNN) architecture that solves the vanishing gradient problem in typical RNNs. It is capable of long-distance, context-dependent learning and storing contextual history. The LSTM is made up of three gate structures such as input gates, forget gates, and output gates. These gates read, write, and reset the network memory unit's state (Ariwa et al., n.d.).

3.6. Experimental setup of ML models

The experimental setting included the full process, from dataset acquisition to ML model refinement and performance assessment, including data preparation, method selection, model efficiency, and final evaluation. In this phase, we will discuss various ML models' experimental setups.[b]

We used several ml classifiers such as DT, SVM, LR, KNN, and NB to predict student choices in their career. Each model was trained using 80% of the dataset, while the remaining 20% was held for testing. The test data coupled with the scoring method was used to determine the accuracy of the prediction of student career choices made by the DT, SVM, and LR models. Also, for SVM, we used a linear kernel function and allowed for probability estimates for each class, making it easier to report prediction confidence levels. Repeatability was guaranteed by using a random seed of 10 for LR. For KNN, the number of neighbors was set to 3 and for NB, we used a Multinomial Naive Bayes classifier. It is important to note that all ML models utilized a uniform strategy to train, forecast, and evaluate the accuracy, facilitating a systematic comparison of their performance in predicting students' career pathways.

For achieving efficient student CP and analyzing data, CNN to predict the data of career, and MLP and LSTM were used. Input data for both training and testing purposes was reshaped to fit each model. Train, they validated it set split: 80:20 on the base of 'stratify' parameter for making it balanced. CNN layers are Convolutional, MaxPooling, Dropout, Flatten, Dense, and Output that are equipped with ReLU as activation. Moreover, for using MLP, ReLU was used with four hidden layers. In case of better performance, they used varying neurons and dropouts to not overfit them. Regarding LSTM, the four used layers, such as LSTM, dense, and two layers, and the dropout was used to regulate this model. All models went through 50 epochs of iteration over the complete training dataset to update weights and reduce gaps between predicted and actual target values.

## 4. Expremental analysis

In this section, we will conduct a comparative discussion of the models' accuracy, precision, recall, and F1 scores. Here we will also discuss the training, testing, and validation accuracy of neural networking models.

4.1. Evaluation matrix of the models

Accuracy is the fraction of correct classifications compared to the total number of instances and is determined by the confusion matrix (Almarabeh, 2017). The formula for accuracy is as follows:

From the above equation (1), we can see the calculation of accuracy.

$$Accuracy = \frac{TN + TP}{TN + FP + FN + FP} \ldots \ldots \ldots (1)$$

---

[b] All source code: https://github.com/sakir101/Career-prediction-main/tree/main

Table 6: Model accuracy percentage

| Classifier Name | Accuracy |
|:---:|:---:|
| DT | 77.27% |
| SVM | 88.63% |
| LR | 84.09% |
| KNN | 77.27% |
| NB | 79.55% |
| CNN | 77.27% |
| MLP | 84.09% |
| LSTM | 84.09% |

From the above Table 6 it is noticeable that SVM has the highest percentage 88.63%. MLP, LSTM and LR have also a good accuracy compared with MLP of over 80%. DT and KNN have the same accuracy at 77.27%. NB and CNN have an average accuracy of 79.55%. The DT achieved an accuracy of 77.27% but may struggle with complex data relationships or overfitting issues. CNN, a deep learning classifier, is less accurate in large datasets. KNN, a simple algorithm, also achieved an accuracy of 77.27% but may be sensitive to noise and outliers. Other classifiers like SVM, LR, NB, MLP, and LSTM achieved higher accuracies.

Precision Refers to the percentage of correctly classified cases from those presented as positive (Almarabeh, 2017). The precision equation (2) is given below:

$$Precision = \frac{TP}{TP + FP} \ldots \ldots \ldots (2)$$

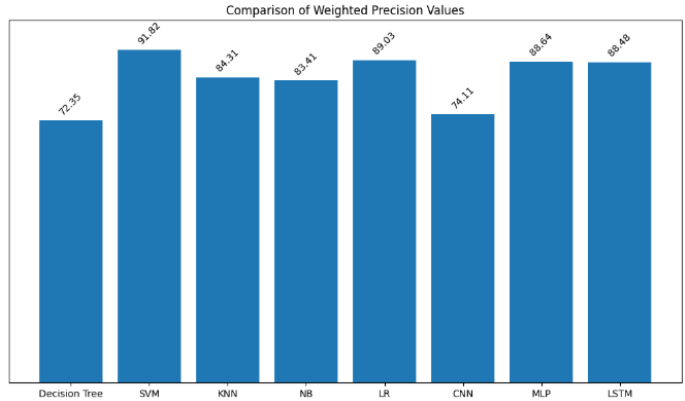

Figure 3: Model precision percentage

SVM has the highest precision rate (91.82%) whereas DT has the lowest precision rate (72.35%). MLP, LSTM, and LR have also good rates of precision compared with SVM respectively 88.64%, 88.48%, and 89.03% as depicted in Fig. 3.

Recall is the fraction of correctly classified positive cases. The equation (3) of recall is given below:

$$Recall = \frac{TP}{TP+FN} \ldots\ldots\ldots (3)$$

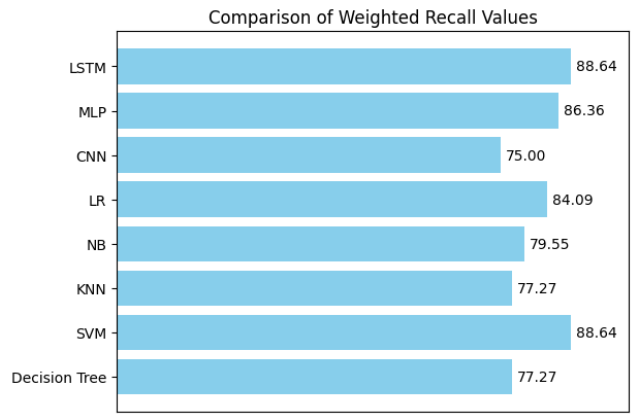

Figure 4: Model recall percentage

As it is seen SVM and LSTM have the similar highest recall percentage (88.64%). MLP and LR also have good recall percentages compared with MLP respectively 86.36% and 84.09% as shown in Fig. 4.

F-measure is a retrieval measure derived from TP, FP, FN, recall, or accuracy values. we calculated the F-measure by using the precision and recall values of the models shown in Eq. (4).

$$F - measure = \frac{2 * Precision * Recall}{Precision + Recall} \quad \ldots\ldots\ldots(4)$$

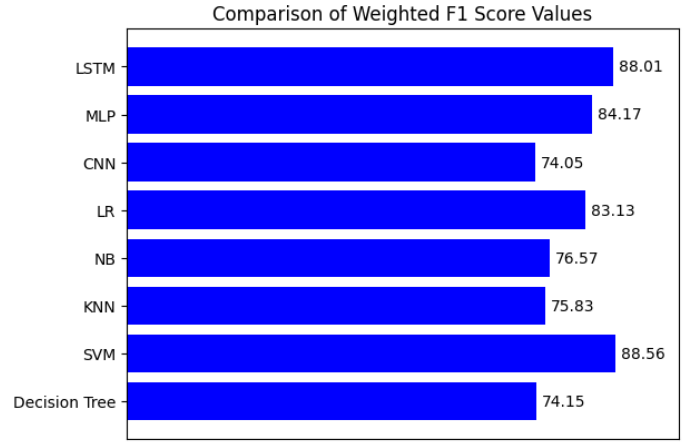

Figure 5: Model f-measure percentage

As can be seen, SVM has the highest F1 score which is 88.56% and LSTM has little bit lower percentage 88.01% as shown in Fig. 5.

4.2. Confusion matrix

The confusion matrix is a thorough depiction of the classifier's performance. This technique helps in assessing multi-class, single-label classification models, where every point of data may only belong to one class at a given moment (Sueno, 2020). In Table 7, we have presented the values for True Positive (TP), False Positive (FP), True Negative (TN), and False Negative (FN) obtained from the confusion matrix of each classifier.

Table 7: Confusion matrix showing model performance in identifying instances

| Classifier Name | TP | FP | TN | FN |
|---|---|---|---|---|
| DT | 34 | 10 | 210 | 8 |
| SVM | 39 | 5 | 215 | 5 |
| LR | 37 | 7 | 213 | 7 |
| KNN | 34 | 10 | 210 | 10 |
| NB | 35 | 9 | 211 | 9 |
| CNN | 33 | 11 | 209 | 11 |
| MLP | 38 | 6 | 214 | 6 |
| LSTM | 39 | 5 | 215 | 5 |

The DT classifier performed reasonably average, with a TP of 34 and a comparable FP count of 10, indicating a reasonable level of accuracy. The SVM and LSTM have the greatest TP count of 39 of all classifiers, with a low FP of 5, indicating their efficacy in properly recognizing positive cases. LR revealed a balanced TP and FP count, indicating good classification accuracy. KNN performed similarly to the DT, with TP and FP counts of 34 and 10, respectively. NB performed similarly, with a TP of 35 and a slightly higher FP of 9. Both the CNN and MLP models have balanced TP and FP counts, indicating their efficacy in classification tasks. Overall, each classifier performed differently in categorizing cases, with SVM and LSTM yielding especially encouraging results in terms of TP and FP counts.

4.3. Macro and weighted average

In Table 8 we showed macro and weighted averages for classifiers

Table 8: Macro and weighted average of the classifiers

| Classifier Name | Macro Average | | | | Weighted Average | | | |
|---|---|---|---|---|---|---|---|---|
| | Precision | Recall | F1-Score | Support | Precision | Recall | F1-Score | Support |
| DT | 0.58 | 0.62 | 0.59 | 44 | 0.72 | 0.77 | 0.74 | 44 |
| SVM | 0.93 | 0.86 | 0.88 | 44 | 0.92 | 0.89 | 0.89 | 44 |
| LR | 0.92 | 0.82 | 0.83 | 44 | 0.89 | 0.84 | 0.83 | 44 |
| KNN | 0.85 | 0.74 | 0.73 | 44 | 0.84 | 0.77 | 0.76 | 44 |
| NB | 0.86 | 0.75 | 0.75 | 44 | 0.83 | 0.80 | 0.77 | 44 |
| CNN | 0.78 | 0.79 | 0.78 | 44 | 0.80 | 0.80 | 0.79 | 44 |
| MLP | 0.94 | 0.88 | 0.88 | 44 | 0.94 | 0.91 | 0.90 | 44 |
| LSTM | 0.79 | 0.79 | 0.78 | 44 | 0.81 | 0.82 | 0.81 | 44 |

4.4. Comparative analysis of deep learning models

In this research we used three deep learning classifiers such as CNN, MLP and LSTM. In this section, we will show a comparative discussion of the three deep learning classifiers based on the learning curves. The learning curve calculates test error based on the training sample count for a classification model and technique (Hoiem et al., 2021).

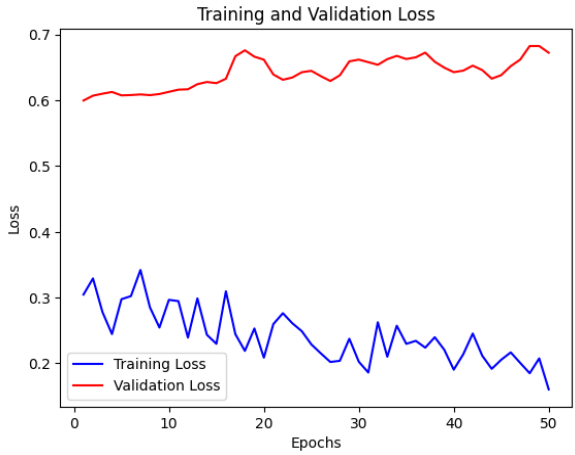

Figure 6: Training and validation loss of CNN

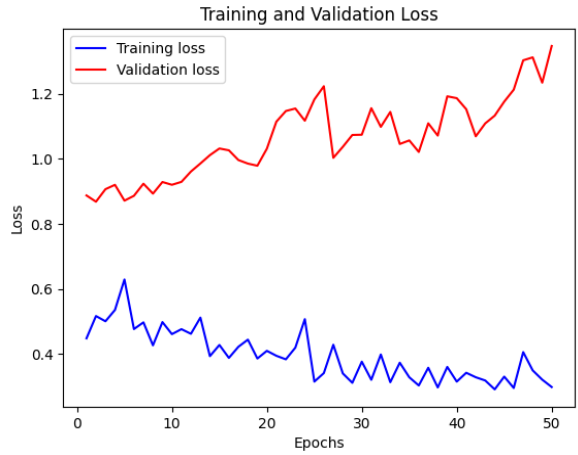

Figure 7: Training and validation loss of MLP

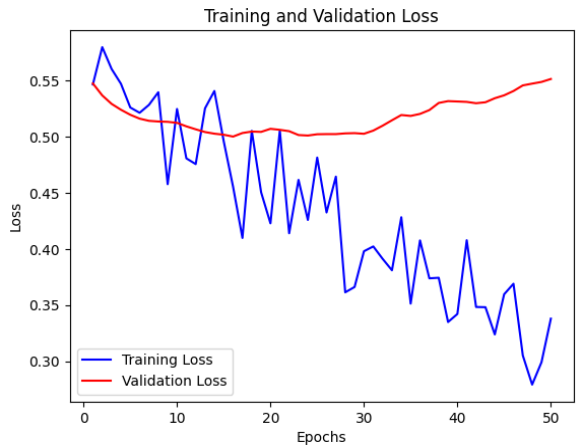

Figure 8: Training and validation loss of LSTM

The CNN, MLP, and LSTM models have different learning curves and performances. The CNN model consistently improves both training and validation accuracies while decreasing training and validation loss. The substantially consistent difference between training and validation accuracies implies good generalization without excessive overfitting as depicted in Fig. 6. In contrast, the MLP model shows quick initial accuracy improvement followed by stabilization, indicating good learning with little overfitting as shown in Fig. 7. Finally, the LSTM model's accuracies grow slowly but steadily, with a clear difference between training and validation accuracies indicating some degree of overfitting, but not severe as shown in Fig. 8. Overall, all models perform rather well, but the CNN model stands out for its consistency and effective generalization. In Table 9 we have shown the training, testing, and validation accuracy of different deep-learning models.

Table 9: Training, testing and validating accuracy of three deep learning classifiers

| Classifier Name | Training Accuracy (Approximate) | Testing Accuracy (Approximate) | Validation Accuracy (Approximate) |
|---|---|---|---|
| CNN | 88.81% | 77.27% | 79.41% |
| MLP | 79.85% | 84.09% | 76.47% |
| LSTM | 79.85% | 84.09% | 85.29% |

The comparison of the three classifiers shows various performance features. The CNN has the best training accuracy (~88.81%), suggesting its capacity to recognize complex patterns in training data. However, it shows somewhat lower testing accuracy (~77.27%) and validation accuracy (~79.41%), indicating mild overfitting. The MLP and LSTM models have comparable training accuracies (~79.85%), somewhat lower than the CNN. The MLP and LSTM models beat the CNN in terms of testing accuracy (~84.09% for both) and validation accuracy (~76.47% for MLP, ~85.29% for LSTM), showing improved generalization ability. While the MLP model has somewhat greater testing accuracy, the LSTM model outperforms the validation set, indicating its ability to capture temporal relationships well. Overall, the LSTM model is the best-balanced alternative, with competitive training accuracy, strong generalization, and good capture of sequential patterns.

## 5. Discussion

Our research utilizes cutting-edge methods to obtain crucial information from students' career objectives and responses to surveys. This technique helps to identify patterns and trends connected with professional preferences and academic interests, providing a better understanding of students' planned destinations. It enables a more detailed understanding of the elements that influence students' decisions and outcomes.

5.1. Discussion on feature engineering concerning survey questions related to student information

Students were surveyed via Google Forms about their professional objectives, chosen fields of study, and necessary skills for their future professions. This data serves as the foundation for the personalized CP system, which provides suggestions based on individual replies. After reviewing the survey results, it is clear that a substantial majority of students are interested in SDE, WD, CS, and AI. These findings highlight the many pathways that CS and SWE students could take.

In this work, we use a hybrid strategy that includes both human and automatic preprocessing procedures, as well as advanced NLP techniques like changing lowercase letters to uppercase, deleting stop words, and vectorizing the text. This comprehensive strategy aims to improve prediction accuracy by eliminating noise and anomalies while retaining the majority of the data. This improves the overall performance of the classifiers.

5.2. Data mining and career prediction

After the data is pre-processed, ML classifiers are used to forecast the future in a variety of fields. Despite the lack of a big dataset, our NLP-based technique incorporates an amazing additional word feature representation, guaranteeing effective text encoding into vectors and resulting in increased accuracy. It is worth noting that there are currently few research in this field, particularly in terms of huge datasets that would allow for successful model training. Despite the dataset's limitations, our research contributes to decreasing the gap by highlighting the promise of NLP techniques for CP.

It is worth noting that prior research, such as (Madhan & Reddy, 2021a) and (Kabakus & Senturk, 2020), did not use NLP techniques to investigate CS and SWE students' careers. Instead, they took traditional ways. This demonstrates the originality of our method, which uses NLP to improve the accuracy of career advising for CS and SWE students. By pioneering the implementation of NLP in this context, our research increases the methodological toolbox available for career predictions, providing new approaches for enhancing the fidelity and relevance of predictions regarding student career guidance.

Our study consists of testing various neural network structures like CNN, MLP, and LSTM models. Each model presents different positives and negatives, with CNN showing a bit of overfitting as well as lower accuracy levels when compared to MLP and LSTM models. However, it is important to mention that our research does not only look into NN models; we also investigate conventional ML methods like DT, SVM, KNN, LR, and NB. Out of these traditional ML models, SVM is found to have the best performance, and LR is not far behind. KNN also gives an impressive performance, but it scores a little less than SVM and LR. NB demonstrates relatively lower performance when compared with LR and KNN, whereas decision tree models are more or less average in terms of performance. Our CP system utilizes the SVM model to predict career paths. Three inputs are analyzed, determining machine-predicted outputs based on our research. This method provides users with percentage-based predictions. So that users can identify their most probable career trajectories.

Table 10: Human giving input and machine-predicted output based on the SVM model

| SL | Input | Machine Predicted Output |
|---|---|---|
| 1 | AI, ML, DS, WD, SDE, GD | SDE: 56.94%, AI: 20.05%, DS: 17.00%, UI / UX: 2.68%, DEV: 1.70%, SEC: 1.63% |
| 2 | ML, DS, CS, ISA | SEC: 52.13%, SDE: 21.38%, AI: 13.99%, DS: 10.78%, UI / UX: 1.41% DEV: 0.32% |
| 3 | AI, ML, DS, DSA, TPA | AI: 49.45%, DS: 40.58%, SDE: 7.98%, SEC: 0.99%, SDE: 0.98%, UI / UX: 0.87%, DEV: 0.15% |

Finally, Table 10 outlines the SVM model's forecast on the optimal career for CS and SWE students based on their skills, interests, and relevant work areas. Output as a percentage allows individuals to easily determine the most likely career path. Overall, our study highlights the reasonable selection and evaluation of the model as an essential aspect of research, considering issues such as dataset properties and complexity. Although the NN models MLP and LSTM display excellent results, investigating the use of classical ML algorithms such as SVM or LR will help to better understand when using a small database. In summation, our study has validated that both NN and traditional ML models are feasible in predicting students' career aspiration. In future studies, we can develop more precise answers for students through NLP improved career predication.

## 6. Conclusion

The CP System will help CS and SWE students choose the right career path and encourage them to grow their ability for the evolving demands of the technology sectors. In this study, we used various classification and deep learning algorithms to try to figure out the most effective approach for CP. The integration of NLP plays a crucial role in the research. It makes it easy to extract valuable insights from the textual data, identify key skills, determine the importance of interests, and understand the context of related works. After analyzing students' data it can be said that most of the students can choose the right career path but they are unable to do so due to lack of proper guidance.

6.1. Future work

In the future, this research will expand its prediction model to include a variety of additional academic departments as well increasing the dataset may give high performance reducing the computational time.

**CRediT authorship contribution statement**

**Sakir Hossain Faruque:** Conceptualization, Investigation, Visualization, Writing - original draft, Data collection, Analysis, Resources, Methodology. **Sharun Akter Khushbu:** Supervision, Data Collection, Writing - review & editing. **Sharmin Akter:** Funding acquisition

**Declaration of competing interest**

The authors declare the following financial interests/personal relationships which may be considered as potential competing interests: Sakir Hossain Faruque reports administrative support and equipment for collecting career-related data from students at Daffodil International University, as well as from students at various other public and private universities in Bangladesh. Sakir Hossain Faruque reports a relationship with Daffodil International University that includes: non-financial support.

In summary, our research based on students' career selection in Bangladesh was conducted professionally, ethically, and with scientific
Integrity. If there are other authors, they declare that they have no known competing financial interests or personal relationships that could have appeared to influence the work reported in this paper.

**Founding Statement**

This research did not receive any specific grant from funding agencies in the public, commercial, or non-profit sectors.

**Acknowledgement**

We express our gratitude to the public and private universities of Bangladesh, especially Daffodil International University, for their assistance in creating the raw datasets.

**Appendix. Explanation of overfitting and underfitting**

In this we will give an overall idea of overfitting and outfitting which are used to evaluate the performance of neural networking models such as CNN, LSTM, and MLP

A.1. Overfitting

Overfitting is a phenomenon wherein a model provides accurate predictions only for the training dataset, but when applied to new data, it produces incorrect predictions. This situation arises for several reasons, such as:
- When the training dataset is not sufficiently large to adequately represent all potential input data values.
- When the training dataset contains irrelevant content.

A.2. Underfitting

Underfitting occurs when a model fails to perform well on the training dataset and consequently does not provide accurate results for new data. Some reasons for underfitting include:
- The model being too simplistic to effectively handle the training dataset.
- Similar to overfitting, if the dataset size is small, it can lead to underfitting.
- Model biases can also contribute to underfitting.

prediction of students career choice. *Procedia Computer Science*, *46*, 176–183. https://doi.org/10.1016/j.procs.2015.02.009

Agarwal, S. (2014). Data mining: Data mining concepts and techniques. *Proceedings - 2013 International Conference on Machine Intelligence Research and Advancement, ICMIRA 2013*, 203–207. https://doi.org/10.1109/ICMIRA.2013.45

Almarabeh, H. (2017). Analysis of Students' Performance by Using Different Data Mining Classifiers. *International Journal of Modern Education and Computer Science*, *9*(8), 9–15. https://doi.org/10.5815/ijmecs.2017.08.02

Alshanik, F., Apon, A., Herzog, A., Safro, I., & Sybrandt, J. (2020). Accelerating Text Mining Using Domain-Specific Stop Word Lists. *Proceedings - 2020 IEEE International Conference on Big Data, Big Data 2020*, 2639–2648. https://doi.org/10.1109/BigData50022.2020.9378226

Ariwa, E., IEEE Technology and Engineering Management Society., & Institute of Electrical and Electronics Engineers. (n.d.). *ICDIM 2018 : Thirteenth International Conference on Digital Information Management (ICDIM 2018) : Berlin, Germany, September 24-26, 2018.*

Casuat, C. D. (2020). Predicting Students' Employability using Support Vector Machine: A SMOTE-Optimized Machine Learning System. *International Journal of Emerging Trends in Engineering Research*, *8*(5), 2101–2106. https://doi.org/10.30534/ijeter/2020/102852020

Hoiem, D., Gupta, T., Li, Z., & Shlapentokh-Rothman, M. M. (2021). *Learning Curves for Analysis of Deep Networks*.

IEEE Staff. (2017). *2017 3rd IEEE International Conference on Computer and Communications (ICCC).* IEEE.

Kabakus, A. T., & Senturk, A. (2020). An analysis of the professional preferences and choices of computer engineering students. *Computer Applications in Engineering Education*, *28*(4), 994–1006. https://doi.org/10.1002/cae.22279

Kamath, C. N., Bukhari, S. S., & Dengel, A. (2018, August 28). Comparative study between traditional machine learning and deep learning approaches for text classification. *Proceedings of the ACM Symposium on Document Engineering 2018, DocEng 2018.* https://doi.org/10.1145/3209280.3209526

Krzeszewska, U., Poniszewska-Marańda, A., & Ochelska-Mierzejewska, J. (2022). Systematic Comparison of Vectorization Methods in Classification Context. *Applied Sciences (Switzerland)*, *12*(10). https://doi.org/10.3390/app12105119

Kumar, D., & Jambheshwar, G. (2011). Performance Analysis of Various Data Mining Algorithms: A Review. In *International Journal of Computer Applications* (Vol. 32, Issue 6).

Lakhotia, S., & Bresson, X. (2018). An experimental comparison of text classification techniques. *Proceedings - 2018 International Conference on Cyberworlds, CW 2018*, 58–65. https://doi.org/10.1109/CW.2018.00022

Lent, R. W., Lopez, A. M., Lopez, F. G., & Sheu, H. Bin. (2008). Social cognitive career theory and the prediction of interests and choice goals in the computing disciplines. *Journal of Vocational Behavior*, *73*(1), 52–62. https://doi.org/10.1016/j.jvb.2008.01.002

Li, Xin., Hu, Wenbin., Wuhan da xue., & Institute of Electrical and Electronics Engineers. (2009). *Proceedings, 2009 International Conference on Information Engineering and Computer Science : ICIECS 2009, Wuhan China 19 -20 December 2009.* IEEE.

Madhan, V., & Reddy, M. (2021a). *Career Prediction System.* https://doi.org/10.32628/IJSRST

Mandalapu, V., & Gong, J. (n.d.). *Studying Factors Influencing the Prediction of Student STEM and Non-STEM Career Choice.*

Muralidhar, K., & Sarathy, R. (2006). Data Shuffling: A New Masking Approach for Numerical Data. In *Source: Management Science* (Vol. 52, Issue 5).

Priyam, A., Gupta, R., Rathee, A., & Srivastava, S. (2013). *Comparative Analysis of Decision Tree Classification Algorithms* (Vol. 3, Issue 2). http://inpressco.com/category/ijcet

Sarker, I. H. (2021). Deep Learning: A Comprehensive Overview on Techniques, Taxonomy, Applications and Research Directions. In *SN Computer Science* (Vol. 2, Issue 6). Springer. https://doi.org/10.1007/s42979-021-00815-1

Shankhdhar, A., Agrawal, A., Sharma, D., Chaturvedi, S., & Pushkarna, M. (2020). Intelligent Decision Support System Using Decision Tree Method for Student Career. *2020 International Conference on Power*